\documentclass[conference]{IEEEtran}
\IEEEoverridecommandlockouts
\usepackage{cite}
\usepackage{subcaption}

\usepackage{amsmath,amssymb,amsfonts}
\usepackage{algorithmic}
\usepackage{graphicx}
\usepackage{textcomp}
\usepackage{xcolor}
\def\BibTeX{{\rm B\kern-.05em{\sc i\kern-.025em b}\kern-.08em
    T\kern-.1667em\lower.7ex\hbox{E}\kern-.125emX}}
    
\usepackage{float}
\floatstyle{plaintop}
\restylefloat{table}
\usepackage{amssymb}
\usepackage{pifont}
\newcommand{\cmark}{\ding{51}}%
\newcommand{\xmark}{\ding{55}}%

\usepackage{booktabs,tabularx}
\usepackage{graphicx}
\usepackage{amsmath}
\usepackage{amssymb}
\usepackage{booktabs}
\newcommand{\Lagr}{\mathcal{L}}
\usepackage{float}
\usepackage[colorinlistoftodos]{todonotes} 
\usepackage{makecell} 

\usepackage[pagebackref,breaklinks,colorlinks]{hyperref}

\usepackage[capitalize]{cleveref}
\crefname{section}{Sec.}{Secs.}
\Crefname{section}{Section}{Sections}
\Crefname{table}{Table}{Tables}
\crefname{table}{Tab.}{Tabs.}
\usepackage{eso-pic}
\newcommand\AtPageUpperMyright[1]{\AtPageUpperLeft{
 \put(\LenToUnit{0.5\paperwidth},\LenToUnit{-1cm}){
     \parbox{0.5\textwidth}{\raggedleft\fontsize{9}{11}\selectfont #1}}
 }}
\newcommand{\conf}[1]{
\AddToShipoutPictureBG*{
\AtPageUpperMyright{#1}
}
}

\conf{International Joint Conference on Neural Networks (IJCNN), 2024} 

\begin{document}

\title{The Bad Batches: Enhancing Self-Supervised Learning in Image Classification Through Representative Batch Curation }

\author{\"Ozg\"u  G\"oksu\\
School of Computing Science\\
University of Glasgow\\
{\tt\small 2718886G@student.gla.ac.uk}
\and
Nicolas Pugeault\\
School of Computing Science\\
University of Glasgow\\
{\tt\small nicolas.pugeault@glasgow.ac.uk}
}
\maketitle

\begin{abstract}
 The pursuit of learning robust representations without human supervision is a longstanding challenge. The recent advancements in self-supervised contrastive learning approaches have demonstrated high performance across various representation learning challenges. However, current methods depend on the random transformation of training examples, resulting in some cases of unrepresentative positive pairs that can have a large impact on learning. This limitation not only impedes the convergence of the learning process but the robustness of the learnt representation as well as requiring larger batch sizes to improve robustness to such bad batches. This paper attempts to alleviate the influence of false positive and false negative pairs by employing pairwise similarity calculations through the Fréchet ResNet Distance (FRD), thereby obtaining robust representations from unlabelled data. The effectiveness of the proposed method is substantiated by empirical results, where a linear classifier trained on self-supervised contrastive representations achieved an impressive 87.74\% top-1 accuracy on STL10 and 99.31\% on the Flower102 dataset. These results emphasize the potential of the proposed approach in pushing the boundaries of the state-of-the-art in self-supervised contrastive learning, particularly for image classification tasks.
\end{abstract}
\section{Introduction}
\label{sec:intro}
\footnotetext{This paper is published in the Proceedings of the International Joint Conference on Neural Networks (IJCNN), with the World Congress in Computational Intelligence (WCCI) 2024.}
Despite the rapid progress of supervised deep learning, the availability of labelled data remains a limitation: transferring the success of deep learning approaches to applications and domains where data and annotations are scarce is a challenge. This has led to the popularity of approaches for unsupervised and self-supervised pre-training: training a deep network to solve a \textit{pretext task} on a large dataset of usually unlabelled examples, before fine-tuning it for the desired downstream task on a typically much smaller, but correctly labelled dataset. This approach has been successful in allowing the use of deep models for tasks where limited labelled data is available. 
One successful example of self-supervised learning is \textit{contrastive learning}, such as SimCLR \cite{simclrv2}, where the pretext task is to optimize a latent space such that augmented versions of the same training example are more similar to each other than to other training examples. 
Contrastive learning has shown excellent performance \cite{dino}, \cite{adco}, \cite{chen2020simple} but requires large datasets, large batch sizes and long training times. 
\begin{figure}[ht]
    \centering
    \includegraphics[scale=0.5]{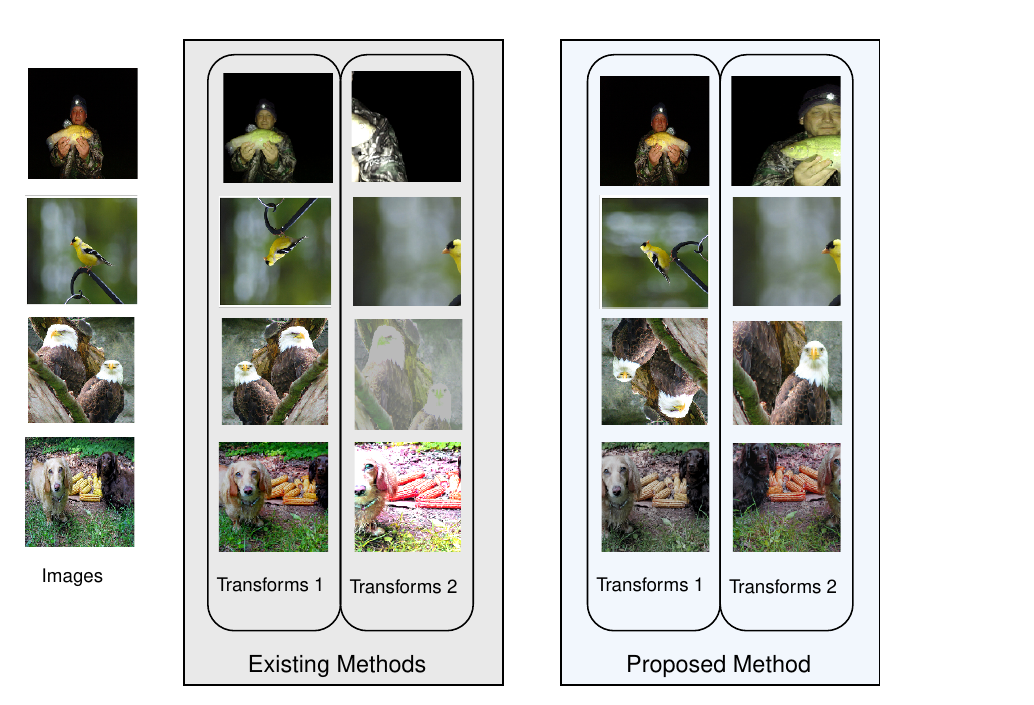}
    \caption{Existing self-supervised contrastive methods mainly rely on various data augmentations to increase diversity, however, it causes weak transformed views of original images. Our method aims to eliminate weak augmented views such as darker images as a similar pair, and insufficient colour changes.}
    \label{fig:problem_def}
\end{figure}
In this article, we propose the hypothesis that contrastive learning is negatively impacted by a relatively small proportion of training examples and augmentations. This is illustrated in Figure~\ref{fig:problem_def}, which shows examples of positive pairs produced by existing methods. In those examples, we observe that data augmentation has degraded essential information in the image, leading to the creation of \textit{false positive}, which could hinder the convergence of the contrastive objective. We argue that discarding such “bad batches” would converge more effectively and efficiently, thereby necessitating smaller batch sizes. 
While self-supervised contrastive learning has shown promise in representation learning, its reliance on randomly-formed batches containing numerous false positives and negatives presents a major obstacle. 
In this work, we present a simple yet effective approach, which involves the evaluation of each batch's representations by measuring their FRD (Fréchet ResNet Distance). FRD can provide a criterion to detect and reject “bad batches” in unsupervised representation learning. We assume that bad batches are characterized by an abundance of false positives and negatives, whereas good batches consist of images with semantically similar views. This study focuses on learning robust representations through the utilization of good batches, which consist of appropriate views of the original images. Contrarily, darker, blurry, or randomly cropped and inappropriate views are intentionally excluded to enhance the quality of the learned representations.
In summary, our major contributions to this work are
\begin{itemize}
    \item a criterion to evaluate augmented batches by Fréchet ResNet Distance (FRD) calculation,
    \item a regularized loss function to reduce the required batch size for training self-supervised contrastive models,
    \item a robust representation learning method by avoiding false positives in batches.
\end{itemize}
This work breaks new ground by proposing a novel approach for evaluating batch quality in self-supervised contrastive learning (SSCL). In traditional SSCL, batches are typically randomly formed, which can lead to the inclusion of distorted or irrelevant image views, thereby introducing false positives and negatives. By implementing a method based on (FRD), we can identify and remove batches that are likely to contain such misleading samples. Our process ensures that only batches with semantically similar views of the original images, referred to as “good batches” are utilized for representation learning. By excluding darker, blurry, or inappropriately cropped views, which are considered irrelevant, the quality of the learned representations is significantly enhanced. This breakthrough has the potential to significantly accelerate the development of efficient and generalizable self-supervised models.
\subsection{Related Work}
Learning representations without human supervision has long been an ongoing challenge due to enhancing generalization and extracting relevant features from raw data.
Self-supervised learning, as opposed to traditional supervised approaches, learns features from unlabelled data and captures representative features as well as the underlying data distribution. Self-supervised learning approaches are divided into two subcategories: generative and discriminative \cite{ozbulak2023know}. 
Generative models such as Generative Adversarial Networks (GANs) and autoencoders are capable of generating realistic, high-quality images from the image dataset. However, they might be insufficient to learn representations, since achieving interpretable representations in the latent space and hard-to-learn informative features are challenging \cite{chen2020simple}. Unlike generative algorithms, discriminative models learn representations through pretext tasks, which focus on understanding the content of images to learn representations instead of generating an image. There are many pretext tasks such as image rotation prediction \cite{johnson2020rotnet}, jigsaw puzzle solving \cite{noroozi2016unsupervised}, image colourization \cite{larsson2017colorization}  and contrastive learning \cite{chen2020simple}, \cite{jaiswal2020survey}, which are designed to replace supervision and extract meaningful and informative representations from raw data, which can later be used for downstream tasks.
Contrastive learning (CL) is a self-supervised learning methodology that is designed to determine a representation space wherein similar instances (positive) are drawn into closeness while dissimilar instances (negative) are pushed apart.  Contrastive learning (CL) facilitates mutual information (MI) among views while preserving task-relevant information. MI evaluates the statistical dependency of two random variables. Regarding CL, the two views of the same image are utilized as random variables. CL attempts to decrease the MI between the two views by maximizing the similarity (and minimizing the dissimilarity) between positive pairs while minimizing the similarity between negative pairs.
The fundamental assumption underlying CL is that obliging the model to obtain representations emphasizing task-relevant information enhances the similarity among positive pairs, facilitating a more straightforward differentiation between positive and negative pairs. This objective is realized by being able to consistently map positive pairs onto proximate points within the feature space, thereby enhancing their ability to distinguish from negative pairs.
Positive pairs encompass images featuring similar views of the original raw data, including various transformations such as rotations or cropping applied to the same image. Conversely, negative pairs are deliberately chosen to be dissimilar, typically involving random combinations from the dataset. Without labels, recent improvements rely on instance discrimination problems in which positive pairs are treated as two augmented versions of the same images while considering the remainder as negatives \cite{chen2020simple}, \cite{moco}. As an illustration, the initial row of the figure \ref{fig:problem_def} shows two transformed views of the same image, captured from distinct angles. CL aims to extract features that are present in both views, like the object's shape and colour. This shared information is generally beneficial for recognizing the object in different contexts. However, some details specific to one angle, like a unique marking or texture, might be discarded as noise during the simplification process in figure \ref{fig:problem_def}. This discarded information could potentially be relevant for a specific task like identifying individual instances of the object. This process ensures that the positive pairs share the same semantic content or meaning. The model can generalize well to downstream tasks by learning to recognize these shared features.
Existing self-supervised contrastive learning methodologies predominantly generate batches through random transformations like cropping and colour jittering. The randomness in batches introduces many weak transformed views (false positives or negatives) and hinders the model's ability to learn from relevant image pairs in figure \ref{fig:problem_def}. Moreover, the requirement to fine-tune transformations for each dataset introduces unnecessary complexity and undermines performance on subsequent tasks.
Several methods eliminate false negative impact on representation, they require larger batch sizes such as 4096 \cite{chen2020simple}, and 8192 \cite{moco}, \cite{dino}. A larger batch provides a more diverse set of negative examples, making it harder for the model to distinguish them based on spurious correlations or artefacts of the data augmentation process. This encourages the model to focus on learning more generalizable features. However, it's important to note that training with large batches requires more memory, which can be a limiting factor for smaller machines or datasets. 
Many approaches to tackle this problem have been offered to eliminate weak data augmentations in batches. GeoDA\cite{cosentino2022geometry} method focuses on shape-preserving geometric augmentations on images and \cite{mishra2021object} proposes context-aware augmentation. \cite{fawzi2016adaptive} aims for effective representation learning by adaptive augmentations based on model uncertainty. The Patch Curation method, as outlined in \cite{welle2021batch}, introduces a batch curation algorithm through the transformation of images into patches. This approach determines batches by assessing their Euclidean distance. Not only, new data augmentation is more specific to use datasets, but also it is not representative.
Previous investigations into batch behaviours have predominantly concentrated on novel data augmentation methods \cite{kurtulucs2023tied}, hard negative mining \cite{zhu2023hnssl}, \cite{robinson2020contrastive}, or the utilization of larger batch sizes for self-supervised contrastive learning. However, batch evaluation by measuring pairwise similarity is not only effective in avoiding weak representation learning on downstream tasks but also benefits from smaller batch size, and sufficient data augmentation tunings. 
Traditional self-supervised learning leverages large-scale data and pretext tasks like image inpainting or contrastive learning to induce informative data representations. While effective, these methods often require significant computational resources. Also, models learn less discriminative features. We propose a novel batch curation approach that significantly improves representation learning efficiency, requiring neither extensive datasets nor large batch sizes. This enables effective self-supervised learning in resource-constrained scenarios.
Larger batches can mitigate random noise in gradients, leading to smoother optimization. However, excessively large sizes can worsen vanishing gradients, hindering efficient network updates. Increased batch size introduces a wider variety of negative examples for contrasting, potentially enriching representations. However, diminishing marginal returns and potential overfitting to sincere correlations within large batches must be balanced. Certain loss functions, like Normalized Temperature-scaled Cross-Entropy (NT-Xent), rely on statistical properties of the data distribution, benefiting from larger batches for accurate estimation. However, over-reliance on larger batches can make performance sensitive to specific data characteristics.
SwAV\cite{caron2020unsupervised} is a self-supervised contrastive learning method, that aims to enhance representation learning by leveraging massive web-based images for data augmentation. BYOL \cite{byol} utilizes a two-network architecture with a student and teacher network without a memory bank or momentum contrast. These models benefit from representation learning on many downstream tasks, however, these models have larger batch size usage. Like SimCLR \cite{chen2020simple} and SimCLR-v2 \cite{simclrv2} train with large batch sizes from 4096 to 16384, SwAV uses even larger batch sizes exceeding 32768. Even BYOL has a simple framework, from 256 to 1024 batch size. However, training larger batches with these models may not consistently be the optimal strategy, and striking the right balance is essential for effective and high-quality representation learning.
SimCLR-v2 aims to improve representation learning by using larger ResNet and inspired memory mechanisms like MoCo \cite{moco} which is a dynamic dictionary that stores past data augmentations or representations. However, they have limitations such as memory management, outdated negative samples in batches, and duplicate samples.
Current methods rely on a pairing strategy that designates two augmented views of the original raw data as positive pairs and randomly selects the remaining transformed views as negative pairs. However, the indiscriminate generation of random images has led to numerous inaccurate augmentations. This has resulted in less discriminative representations and a diminished generalization capacity for self-supervised contrastive models.
These methods focus on representation learning in self-supervised contrastive learning; however, they cannot extract robust representations and have poor performance on downstream tasks. In addition, they might not be efficient in reducing gradient noise, eliminating false positive pairs in batches, and loss function behaviour, which requires a larger batch size due to their reliance on data distributions.

\section{Methodology}
The self-supervised model employs a predetermined method for curating random batches without evaluating individual batches. Fréchet ResNet Distance (FRD) scores for each batch are computed at the nth epoch, before the activation of the batch curation algorithm. The rationale behind applying the batch curation method in the early stages of learning is to leverage the advantages of features acquired during the initial epochs.
As a subsequent step, we seek to enhance the model's robustness to false positives or noise within the batch. To achieve this, we introduce regularization to the contrastive loss by incorporating the Huber loss \cite{barron2019general} with an associated coefficient.
\begin{figure} [htp!]
    \centering
    \includegraphics[width=0.5\textwidth]{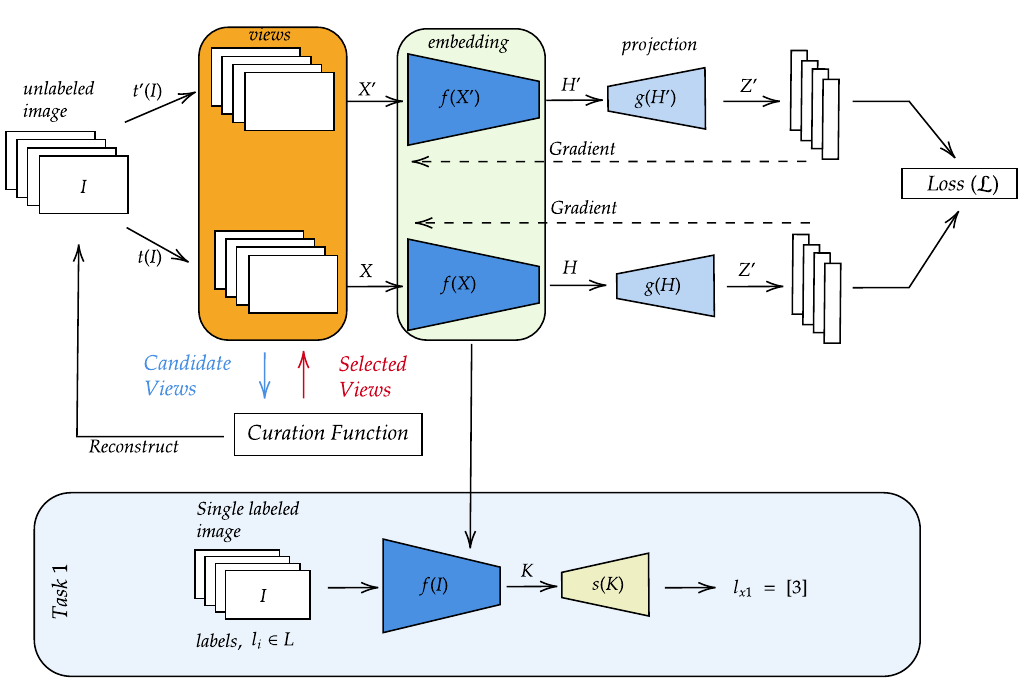}
    \caption{ Our presented framework for batch curation in self-supervised contrastive learning. Task 1 illustrates image classification as a downstream task. The batch curation part mainly decides which batches are used to update gradients. }
    \label{fig:fid}
\end{figure}
\subsection{Contrastive Learning}
Following the contrastive learning framework (as proposed by, e.g., \cite{chen2020simple}), we have a training set of images $x \in \mathcal{D}$, a family of transformations $\mathcal{T}$, a backbone network $f$ such that $h = f(x)$ is a latent encoding of $x$ and a projection head $g$ such that $z = g(z)$ is a projection of $h$, as illustrated in Figure~\ref{fig:fid}. 
Considering a batch of $N$ training examples $B = \left\{x_1, x_2, \ldots, x_N\right\} \subset \mathcal{D}$, by applying randomly sampled transformations $t \in \mathcal{T}$, we generate two augmented batches $B_{1}$ and $B_{2}$, for a total of $2N$ data points. 
We say that two projection vectors $z_i$ and $z_j$ form a \textit{positive pair} if they are transformations of the same training image $I$, a \textit{negative pair} if they are transformations from two distinct images. 
Contrastive learning approaches in the literature predominantly employ the Normalized Temperature-scaled Cross-Entropy (NT-Xent) loss\cite{chen2020simple}, \cite{SiameseIM}. 
It is typical to concatenate the projection vectors from $B_1$ and $B_2$ in a single tensor of $2N$ vectors $z_i$. Then if $z_i,z_j$ is a positive pair of projections, the NT-Xent loss over those $2N$ projections is: 
\begin{equation} \label{eq:4}
    \Lagr_{i, j} = -\log\frac{exp(sim(z_{i}, z_{j})/ \tau)}{\sum_{k = 1}^{2N} {1}_{[k \neq i]} exp(sim(z_{i}, z_{k}) / \tau) }
\end{equation}
The function $sim(.,.)$ represents the cosine similarity, and $2N - 1$ denotes the number of negative (dissimilar) samples since $z_i$, $z_j$ are positive (similar) samples transformed in the same image and $\tau$ is the temperature value.
Because Equation~\ref{eq:eq4} is asymmetric, we calculate the final loss by the loss of a batch $B$ is defined as
\begin{equation}
    \mathcal{L}_{nt \mbox{-}xent}=\frac{1}{2 N} \sum_{k=1}^{N}\left[\Lagr_{2 k-1,2 k}+\Lagr_{2 k, 2 k-1}\right]
    \label{eq:nxent}
\end{equation} 
The contrastive learning literature has shown that minimizing this loss produces a latent space $\mathcal{H}$ that can allow solving many downstream tasks by just training a new projection head, requiring limited sets of labelled examples. 
In this article, we argue that contrastive learning can be severely affected by just a few instances of poor data augmentations. In the following, we describe two approaches to reduce the impact of those: Huber loss regularization and Fréchet batch curation.
\subsection{Huber Loss Regularization} 
Contrastive learning ensures that projections of the same image (positive pairs) are more similar than projections of other samples. We argue that convergence can be improved by ensuring that the projections $z_i$ and $z_j$ for two different data augmentations $t_i, t_j$ of the same image $I$ should be similar. This can be enforced by adding a regularization term to the loss. When there is considerable dissimilarity between positive pairs, the Huber loss imposes a penalty on the loss value, encouraging the model to bring the representations closer together in the latent space. Due to its proven effectiveness in handling outliers, the Huber loss ensures enhanced resilience and stability throughout the learning process.
In practice, for a positive pair $ z_{i} $, $ z_{j} $, we have
\begin{equation}
    \footnotesize
    l_{i,j}= \begin{cases}0.5  \left(z_{i}-z_{j}\right)^{2}, & \text { if }\left|z_{i}-z_{j}\right|<\text {\(\delta\)  } \\ \text { \(\delta\) } \left(\left|z_{i}-z_{j}\right|-0.5  \text { \(\delta\) }\right), & \text { otherwise }\end{cases}.
\end{equation}
where the parameter $\delta$ determines the point at which the loss function transitions from quadratic to absolute, set to 1.0 in our experiments. 
Over the two transformed versions $B_1$ and $B_2$ of a batch $B$ of $N$ images, the Huber loss is defined as the average loss over all $N$ positive pairs in $B_1$ and $B_2$:  
\begin{equation}
\mathcal{L}_{huber} = \operatorname{mean}(l_1, ..., l_N).
\end{equation} 
Including this term, the new regularized objective function becomes
\begin{equation}
    \mathcal{L} = \mathcal{L}_{nt \mbox{-}xent} + \lambda  \mathcal{L}_{huber}
\end{equation}
Notably, our approach distinguishes itself by not necessitating the utilization of a memory bank, a larger batch size, extended training periods for effective generalization, or a huge dataset like ImageNet with a substantial number of samples.
\subsection{Fréchet Batch Curation}
In this paper, we mainly focus on two problems; avoiding learning false positive and negative views, and updating the gradient with each randomly curated batch. In our reasonable non-stochastic method; the FRD score measures the similarity between two distributions of augmented images which is less than the threshold value, the batch updates the gradients otherwise the batch is reconstructed. The threshold value is calculated as the average FRD scores of batches at epoch 5. We calculate the threshold value of batches after a few epochs of training since random initialization of model weights can be harmful to FRD scores. 
During the initial five epochs, we employ algorithms without batch curation to facilitate the learning of representations. In our experiments, empirical testing of various starting epochs reveals that the 5th epoch is optimal as the starting point for curating the batch algorithm.
We hypothesize that some unusual combination of image transformation and training images can produce misleading examples (as illustrated in Figure|\ref{fig:problem_def}) that hampers the convergence and performance of contrastive learning. This section proposes to discard “bad batches” $B$ by considering the average FRD between the two augmented batches $B_{1}$ and $B_{2}$. 
The Fréchet distance calculates the difference between two Normal distributions. In the generative learning field, the Fréchet Inception Distance (FID) is commonly employed to measure the similarity between real and fake images in the generative learning literature. 
We extract representations in the latent space including the projection head, as depicted in figure \ref{fig:fid} ($Z, Z'$), for (FRD) scores.
The Fréchet ResNet Distance $FRD(\cdot)$, inspired by FID \cite{heusel2017gans}, calculates the difference between the augmented views of an image batch as distributions in the latent space of ResNet backbone network. Given a candidate batch $B$, and two augmented batches $B_{1}, B_{2}$ produced by the application of random transformations drawn from $\mathcal{T}$, we can estimate the mean and covariance $\mu_{1}$, $\mu_{2}$ and $\Sigma_{1}$, $ \Sigma_{2}$ of their latent ResNet representations, and calculate the FRD: 
\begin{equation}\thead{
    \mathrm{FRD (B_{1}, B_{2})}=\| \mu_{1}-\mu_{2}\|^{2}+ \\ \newline \operatorname{tr}\left(\Sigma_{1}+\Sigma_{2}-2\left(\Sigma_{1} \Sigma_{2}\right)^{1 / 2}\right)}
\end{equation}
We then propose as threshold the average FRD distance between augmented batches $B_{k1}, B_{k2}$ over all batches $B_{k}$ in the dataset $\mathcal{D}$
\begin{equation}
    \tau_{FRD} =\frac{1}{M} \sum_{k=1}^{M}\left( FRD(B_{k1}, B_{k2})\right )
\end{equation}
where $M$ is the number of batches in the dataset, $B_{k1}$ and  $B_{k2}$ are augmented views of $B_k$ batch. 
Then proposed FRD batch curation discards a candidate batch $B_k$ if $\mathrm{FRD (B_{k1}, B_{k2})} > \tau_{FRD}$.
Our presented batch curation framework in figure \ref{fig:fid} for self-supervised learning follows default parameter training for the first 5 epochs. At the fifth epoch, we calculated the average FRD scores of each batch as a threshold value for the following epochs. Using an average threshold value for batch evaluation can provide balance in data distributions instead of using maximum or minimum in a dataset. The threshold value is 0.56 for our experiments. The FRD score shows the similarity of views, and smaller scores represent highly similar views; otherwise, data augmentation can transform hardly similar views. The process uses similar, better-augmented views as a good batch for representation learning. In this study, the average value, referred to as the threshold, was employed in the evaluation of various representations. The rationale behind utilizing trained representations for threshold calculation is grounded in the pursuit of obtaining quantifiable and representative features. This ensures that the threshold, a pivotal parameter in the evaluation process, is derived from a set of features that are not only informative but also reflective of the model's learning capabilities. By focusing on the 5th epoch, which has been identified as the epoch of peak performance, we aim to capture and leverage the most effective and discriminative features embedded in the trained representations.
In essence, this approach seeks to enhance the interpretability and reliability of the threshold by basing it on features extracted from representations at a specific epoch that has proven to yield optimal results.

\section{Evaluation}
To validate our hypotheses and the proposed approach, we compared performance with baseline contrastive learning approaches on multiple datasets. 
Moreover, we systematically explore the impact of the Huber loss, akin to $L1$ and $L2$ losses, by conducting linear evaluations and transfer learning assessments. These evaluations are carried out across widely recognized datasets, providing a comprehensive investigation into the performance and capabilities of our proposed approach relative to existing methods. 

\textbf{Datasets and Metrics:} Self-supervised approaches generally use ImageNet for pre-training with large batch sizes such as 4096 or 8192 to improve robustness to outliers at the cost of memory. We argue that our improved approach can perform with smaller batch sizes and datasets. Accordingly, we tested models trained on ImageNet and CIFAR10 for self-supervised learning and tested the learnt representations on a range of downstream tasks such as CIFAR100 \cite{chen2020simple} (resolution $32 \times 32$, 100 classes), STL10 \cite{coates2011analysis} (resolution $96 \times 96$, 10 classes), ImageNet\cite{deng2009imagenet} (resolution $224 \times 224$, 1000 classes), Caltech101 \cite{fei2004learning} (resolution $224 \times 224$, 101 classes), Flower102 \cite{nilsback2008automated} (resolution $ 224\times 224 $, 102 classes), and MNIST \cite{deng2012mnist} (resolution $28 \times 28$, 10 classes). 

\textbf{Small-scale Training:}  We use random cropping and resizing, colour jittering, random horizontal flipping, and random greyscale for CIFAR10 as transformation to create pairs. We use a Resnet-50 architecture with a smaller convolution kernel size (3 $\times$ 3) as a base encoder and a non-linear projection head with two linear layers with batch normalization to project representations to a 128-dimensional latent space. Furthermore, we use a lambda warm-up scheduler for 30 epochs, and following that we continue the cosine decay scheduler without restarting for training. The only difference between CIFAR10 and ImageNet is that for the latter, the first layer of the encoder network has kernel size (7 $\times$ 7), to ensure that the latent space is of the same dimension. The choice of different kernel sizes for extracting representative features is motivated by the disparity in resolution between ImageNet and CIFAR-10 datasets.

\textbf{Loss Function and Batch Size:} We enhance the contrastive loss with a regularizing Huber loss, with a $\delta$ parameter to tune the impact of sensitivity. Outliers can cause significant errors, and Huber loss can address the error problem by transitioning from a quadratic loss which is similar to $L2$ loss for small errors and absolute, $L1$ loss for errors. In our case, we assume that false positives and negatives can be considered outliers and noise within the representation space. The $\delta$ parameter, which is set to 1.0 in our experiments, can control these changes. 
\section{Results}
\begin{figure}[ht]
    \centering
    \begin{subfigure}{0.15\textwidth}
        \centering
        \includegraphics[width=\linewidth]{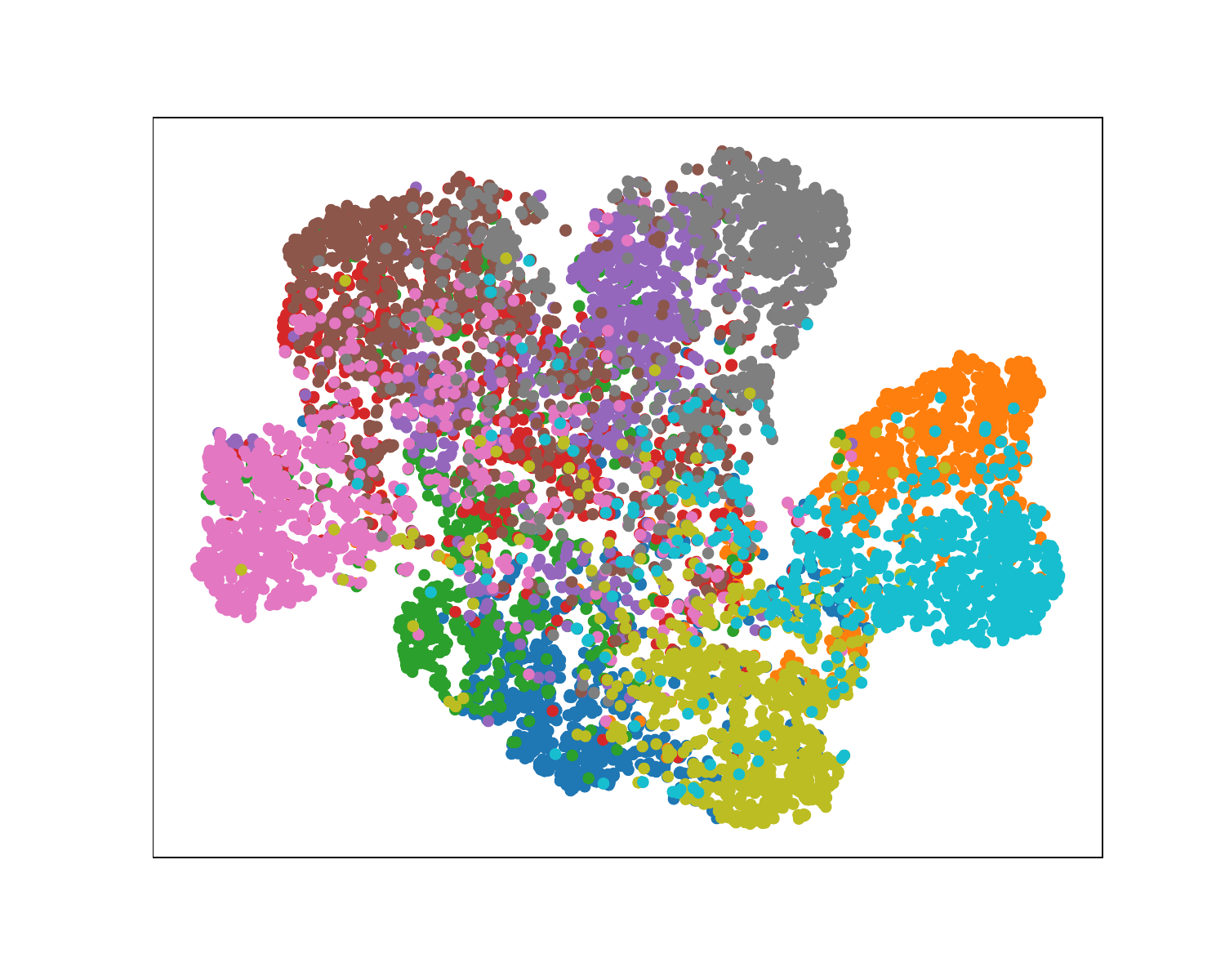}
        \caption{SimCLR (C) }
        \label{fig:sub1}
    \end{subfigure}%
    \begin{subfigure}{0.15\textwidth}
        \centering
        \includegraphics[width=\linewidth]{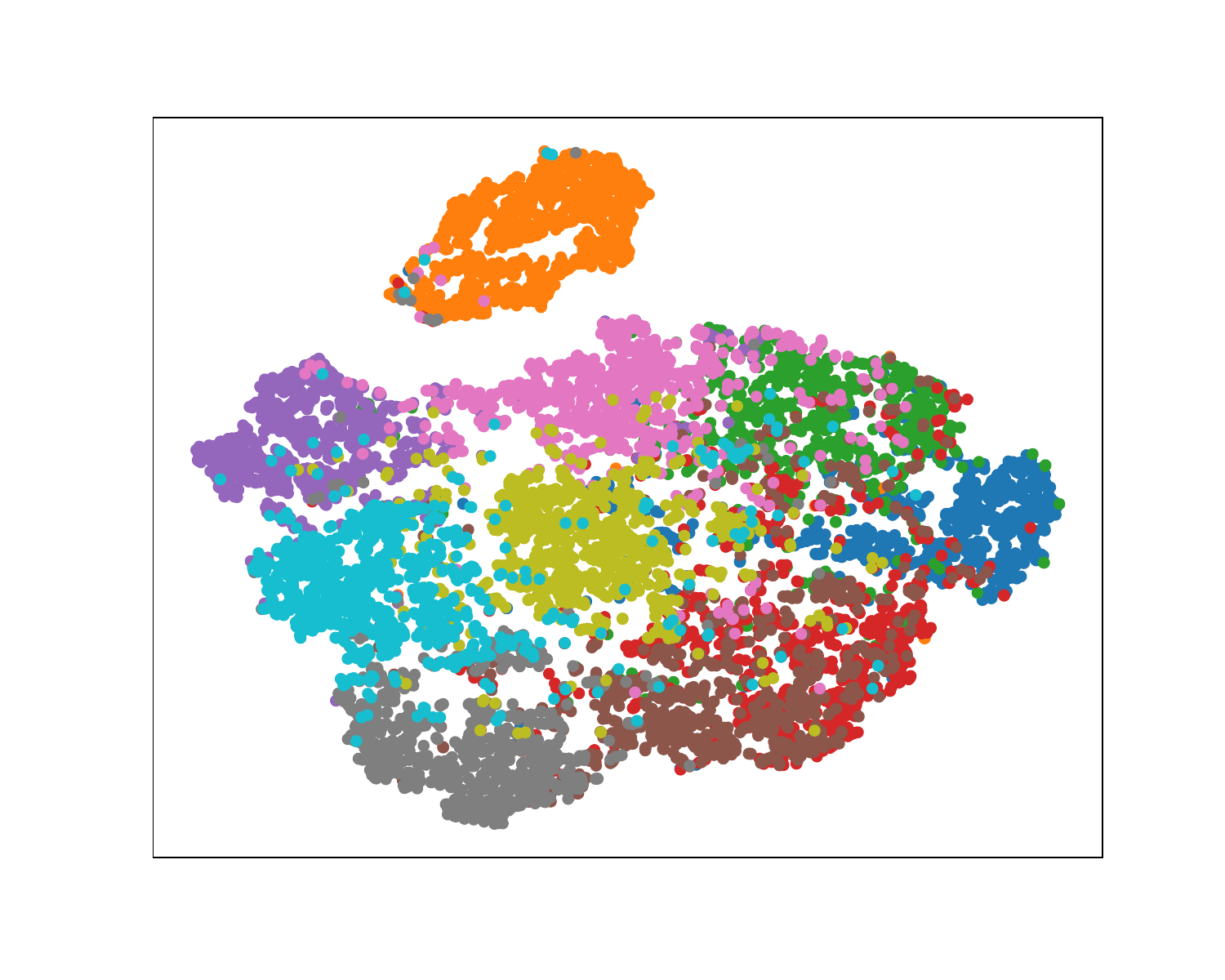}
        \caption{SimCLR (M) }
        \label{fig:sub2}
    \end{subfigure}%
    \begin{subfigure}{0.15\textwidth}
        \centering
        \includegraphics[width=\linewidth]{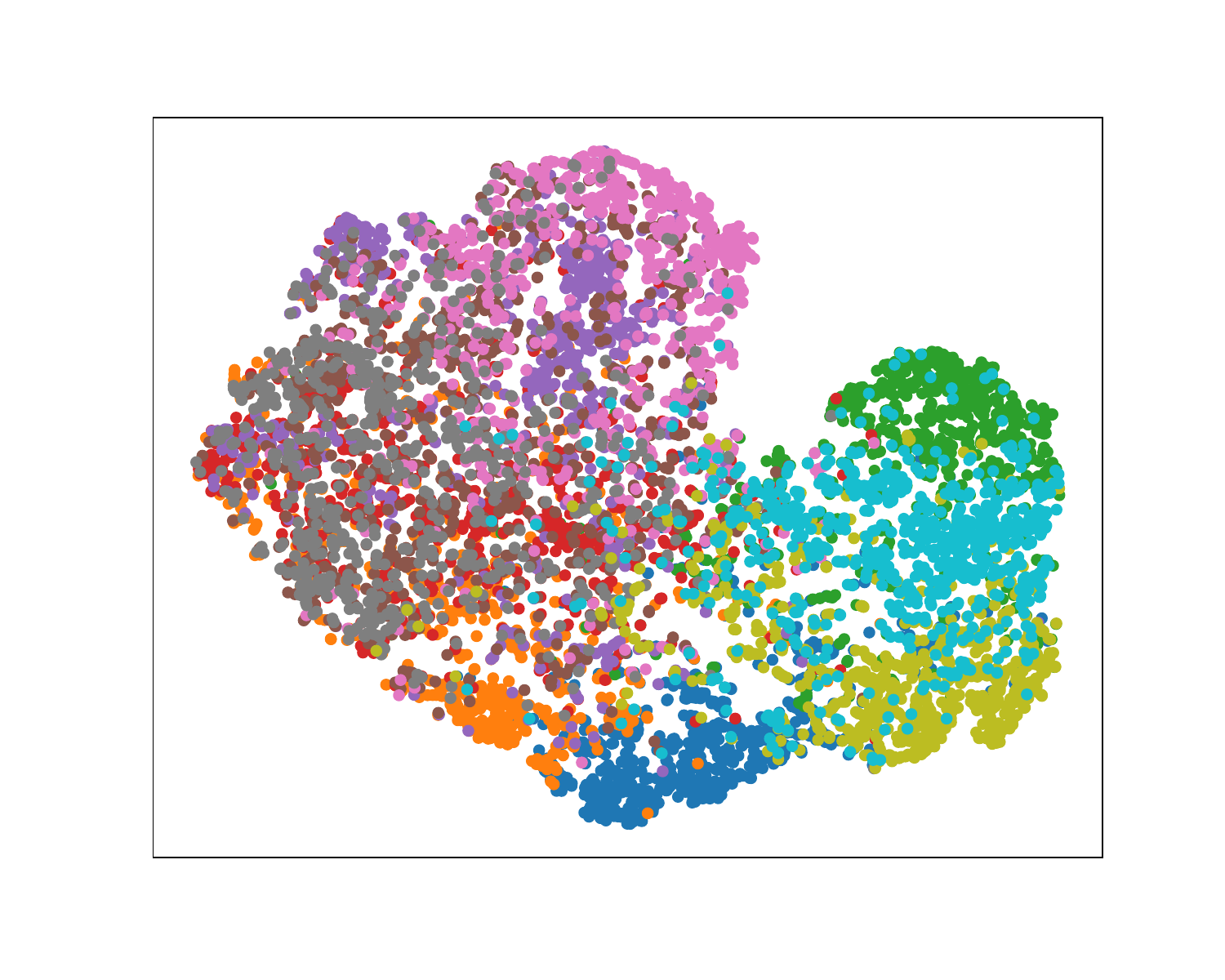}
        \caption{SimCLR (S) }
        \label{fig:sub3}
    \end{subfigure}
    
    \vspace{1em}

    \begin{subfigure}{0.15\textwidth}
        \centering
        \includegraphics[width=\linewidth]{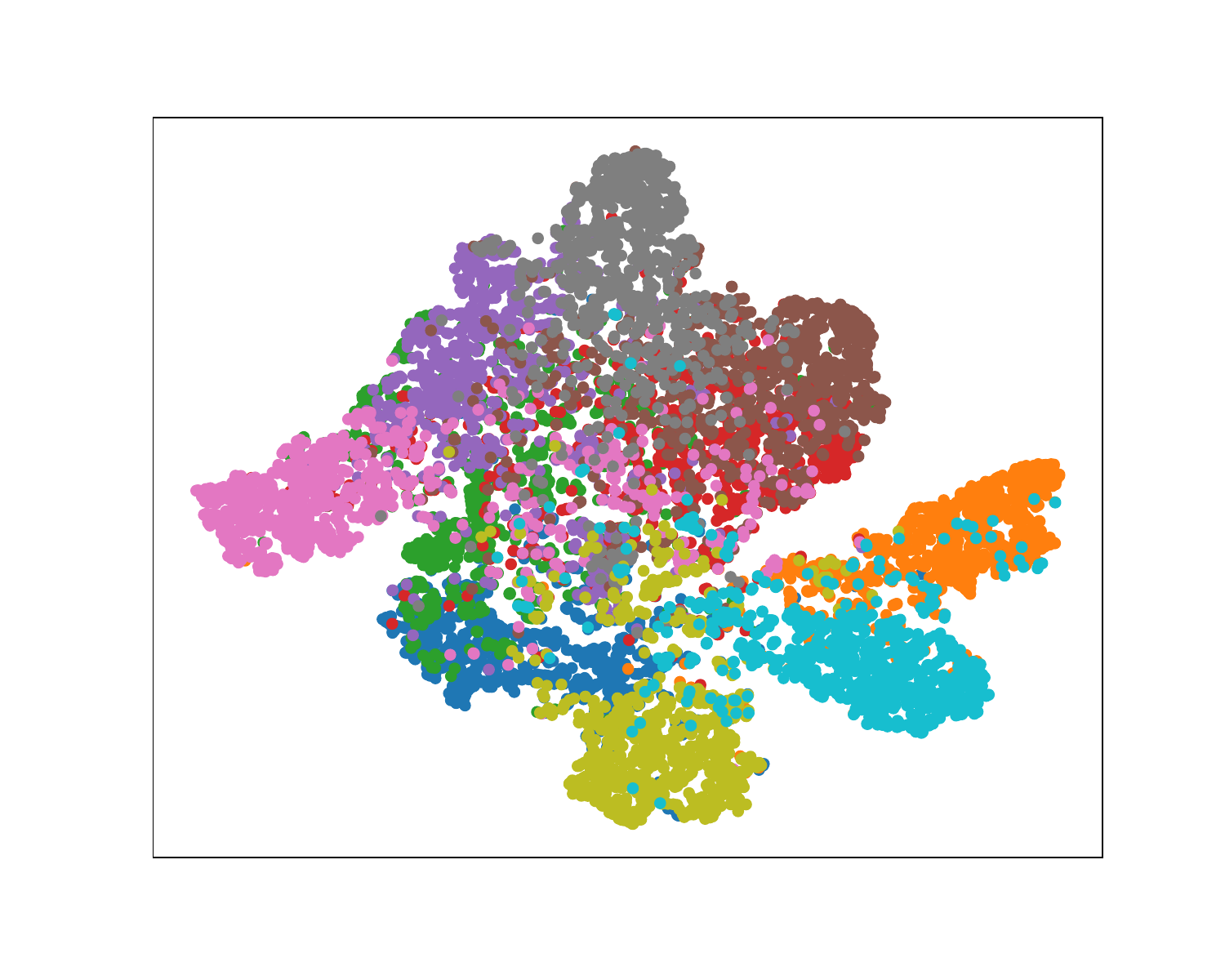}
        \caption{Ours-H (C)}
        \label{fig:sub4}
    \end{subfigure}%
    \begin{subfigure}{0.15\textwidth}
        \centering
        \includegraphics[width=\linewidth]{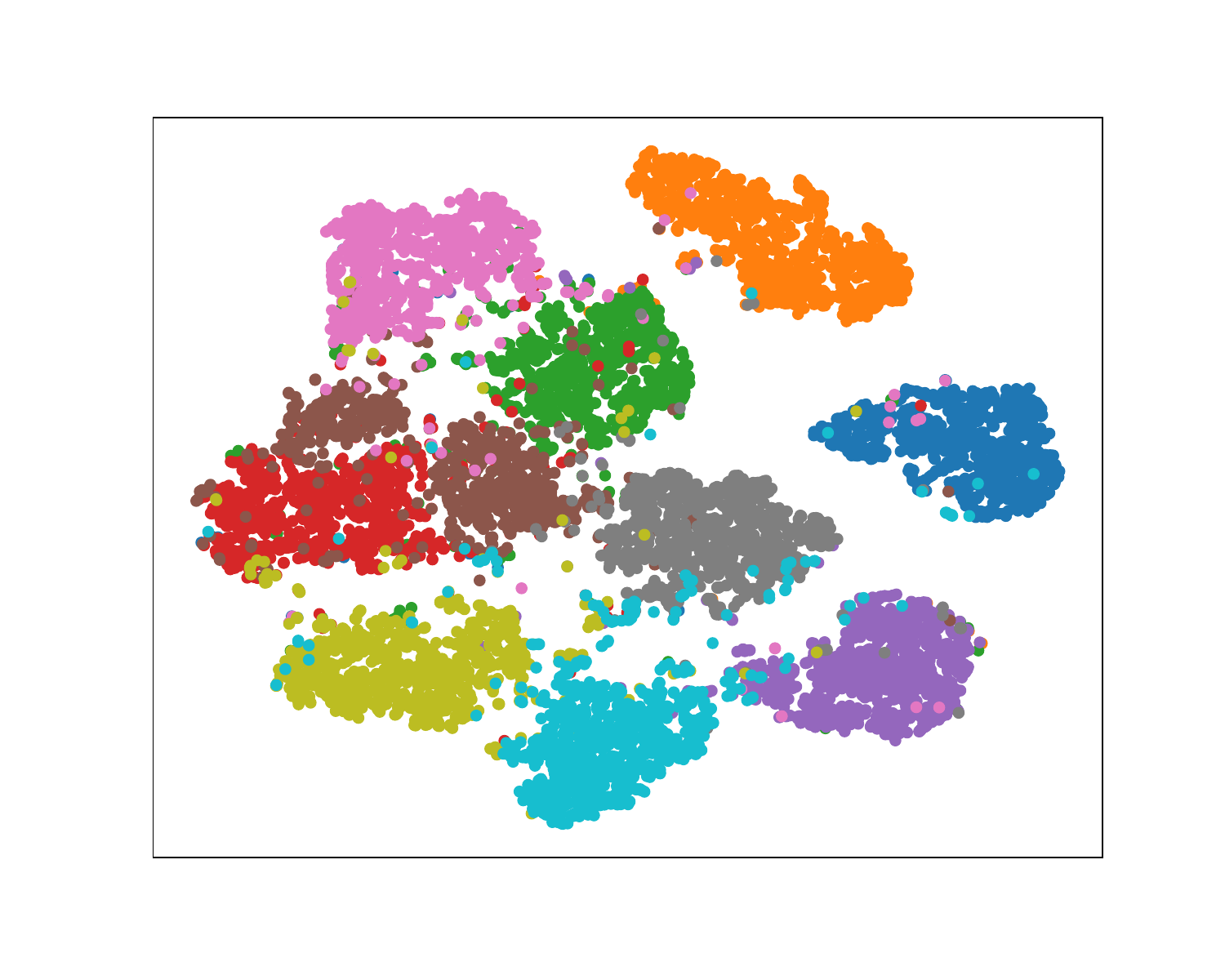}
        \caption{Ours-H (M)}
        \label{fig:sub5}
    \end{subfigure}%
    \begin{subfigure}{0.15\textwidth}
        \centering
        \includegraphics[width=\linewidth]{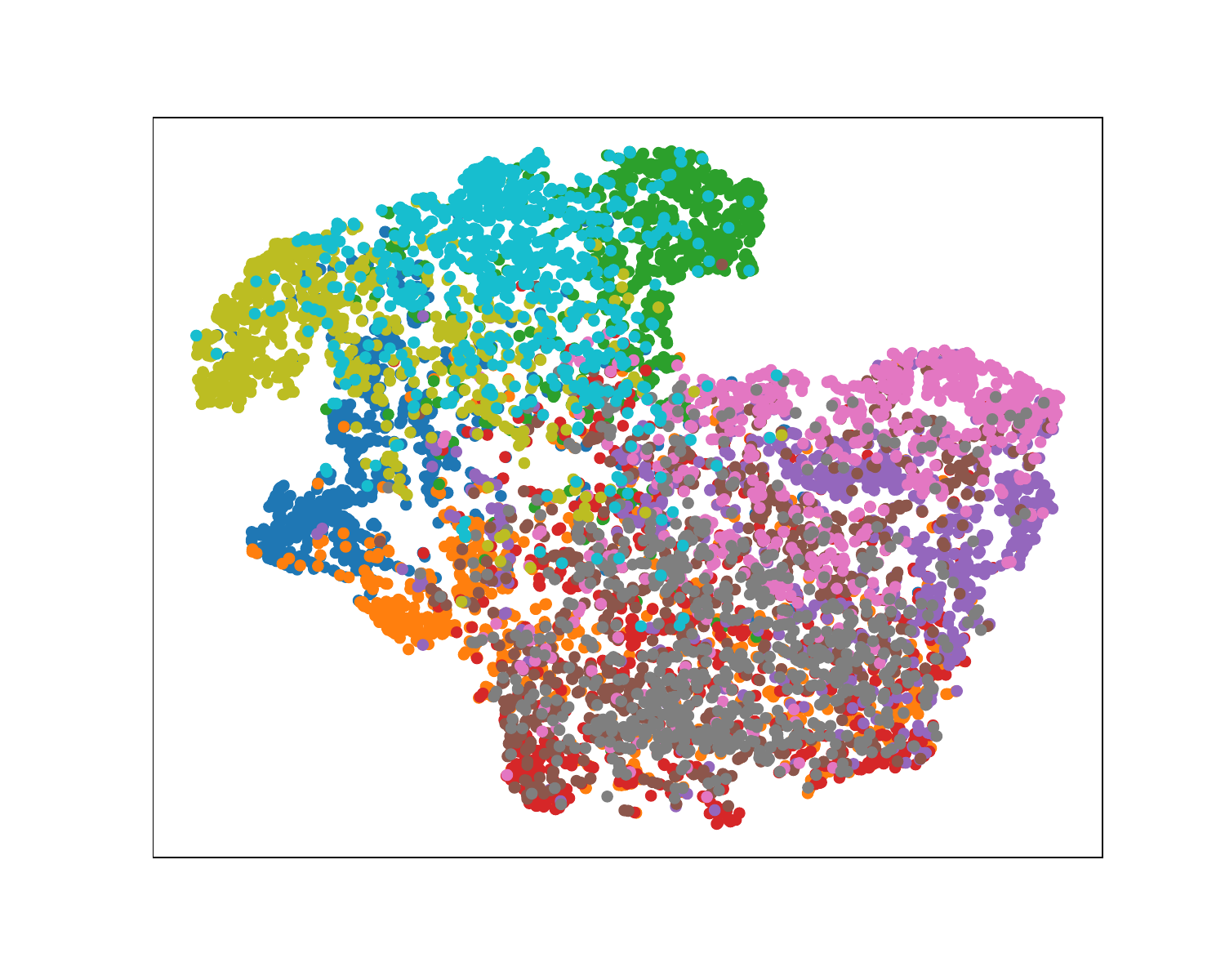}
        \caption{Ours-H (S)}
        \label{fig:sub6}
    \end{subfigure}
    
    \vspace{1em}

    \begin{subfigure}{0.15\textwidth}
        \centering
        \includegraphics[width=\linewidth]{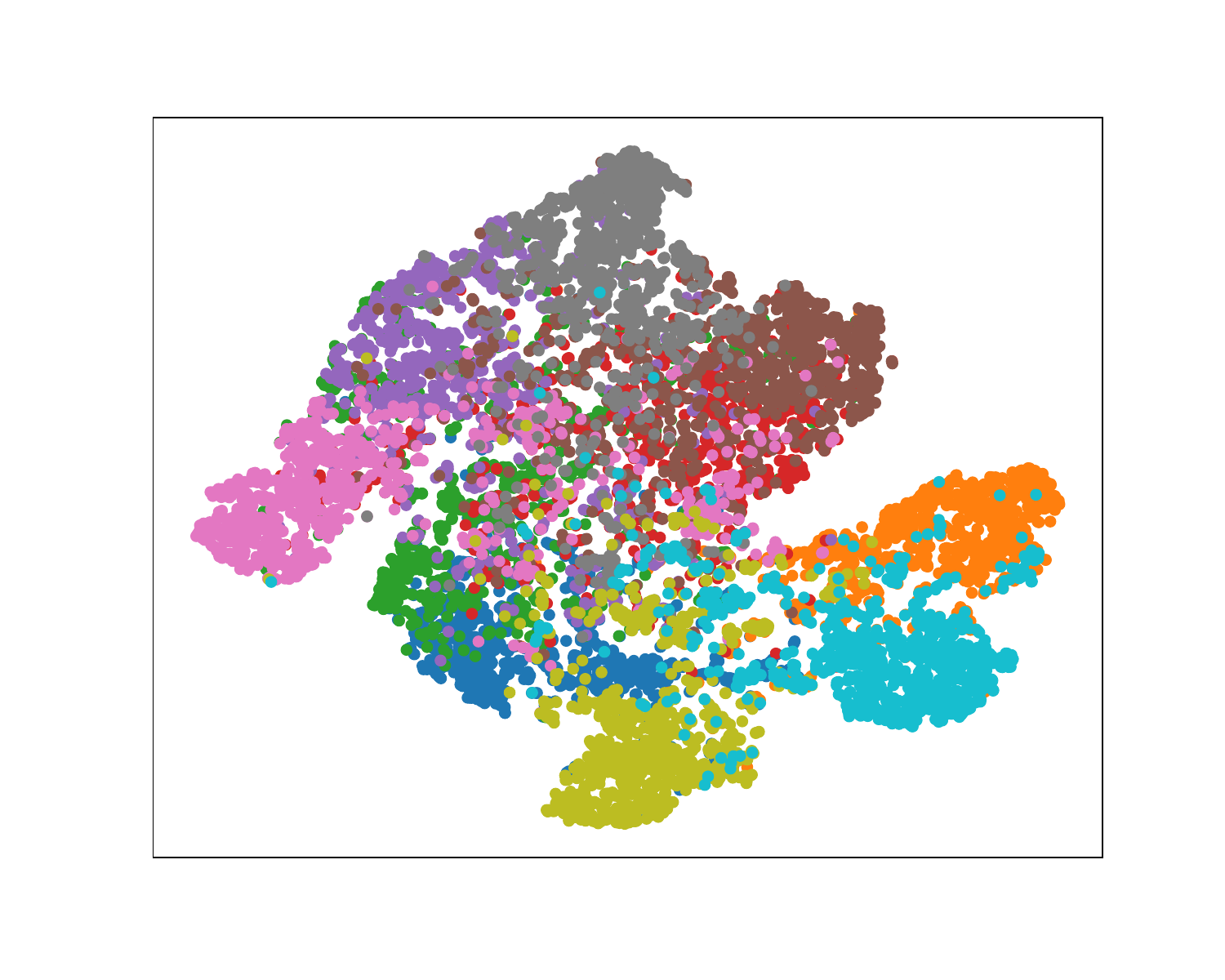}
        \caption{Ours-F (C)}
        \label{fig:sub7}
    \end{subfigure}%
    \begin{subfigure}{0.15\textwidth}
        \centering
        \includegraphics[width=\linewidth]{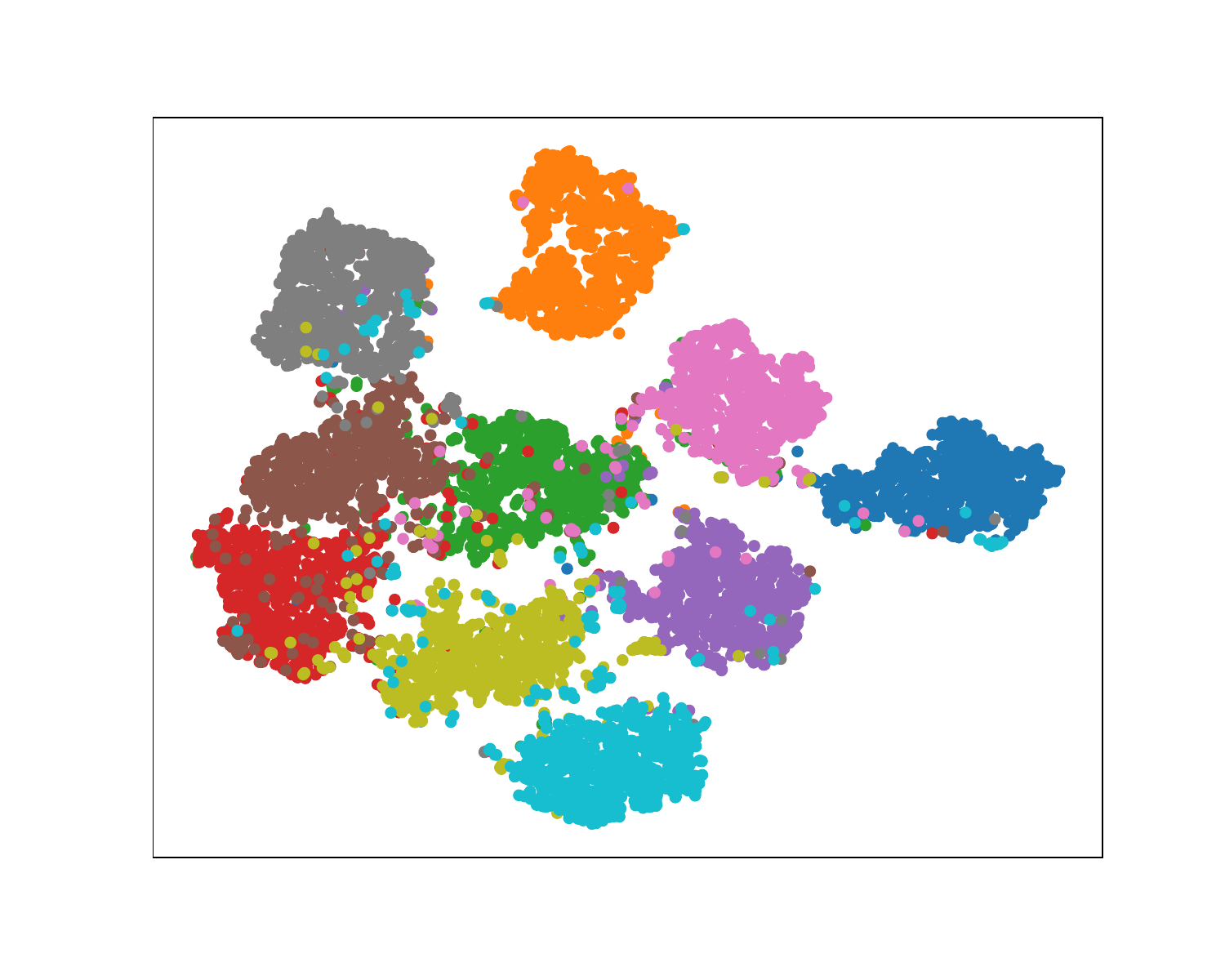}
        \caption{Ours-F (M)}
        \label{fig:sub8}
    \end{subfigure}%
    \begin{subfigure}{0.15\textwidth}
        \centering
        \includegraphics[width=\linewidth]{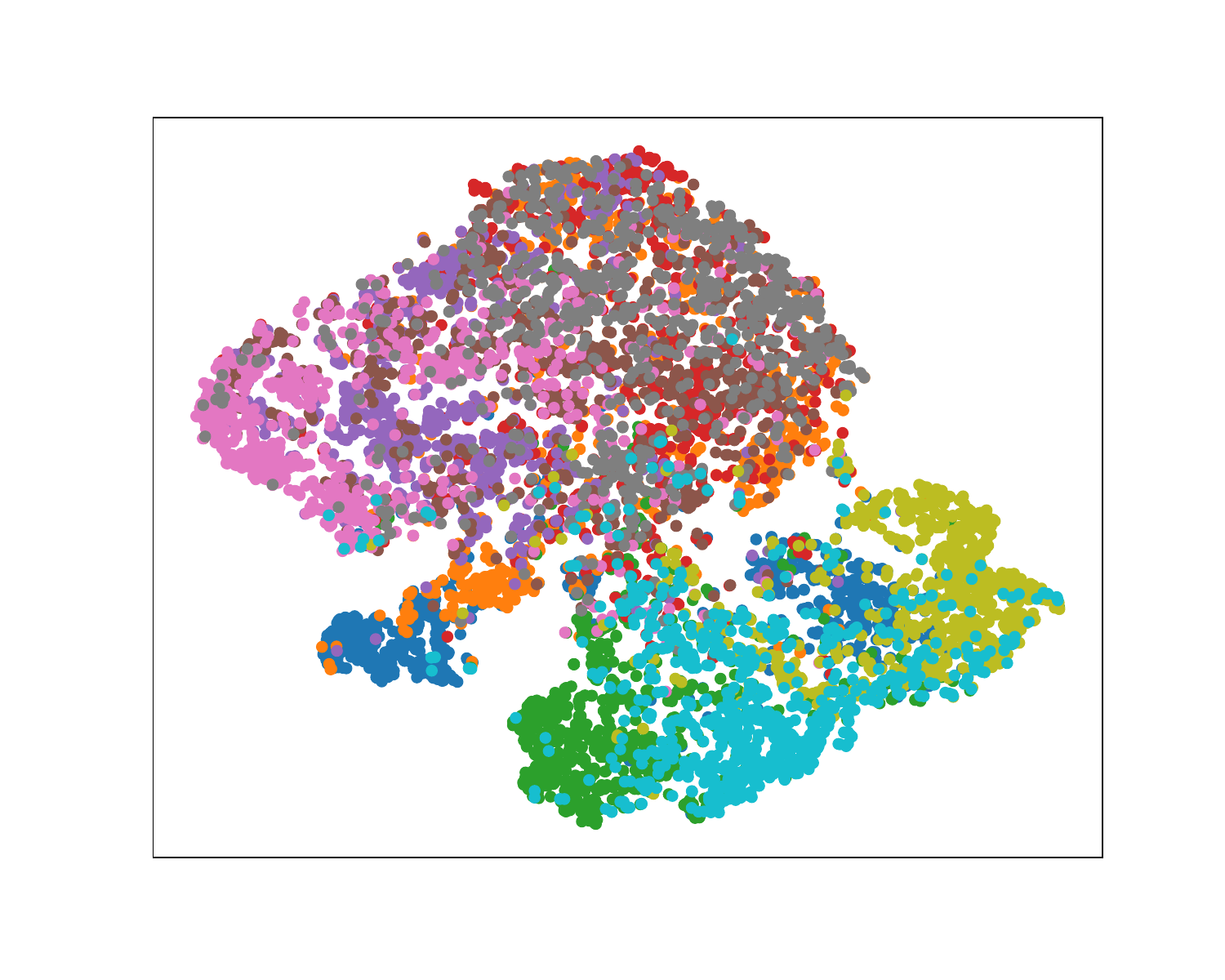}
        \caption{Ours-F (S)}
        \label{fig:sub9}
    \end{subfigure}
    
    \vspace{1em}

    \begin{subfigure}{0.15\textwidth}
        \centering
        \includegraphics[width=\linewidth]{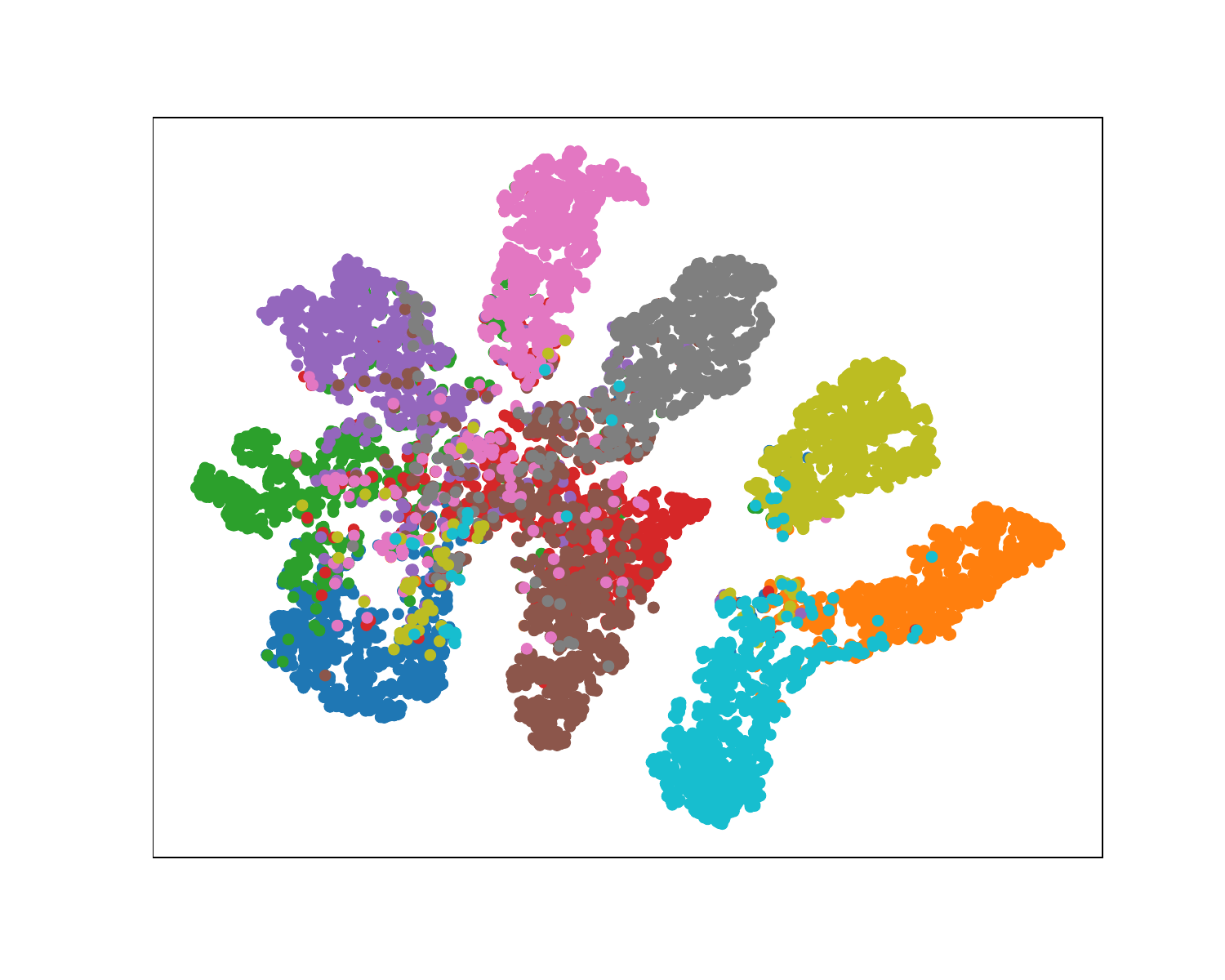}
        \caption{Ours (C)}
        \label{fig:sub10}
    \end{subfigure}%
    \begin{subfigure}{0.15\textwidth}
        \centering
        \includegraphics[width=\linewidth]{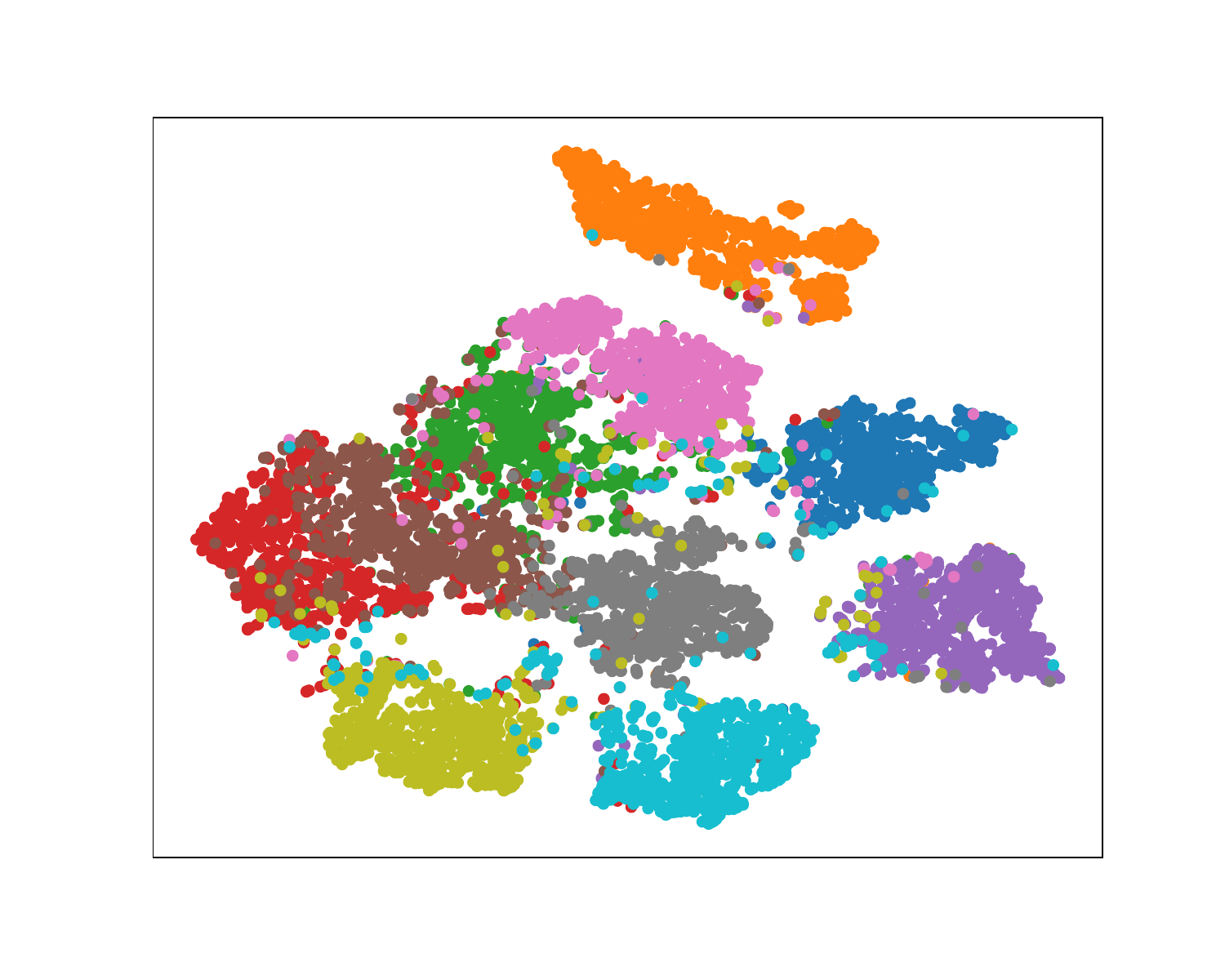}
        \caption{Ours (M)}
        \label{fig:sub11}
    \end{subfigure}%
    \begin{subfigure}{0.15\textwidth}
        \centering
        \includegraphics[width=\linewidth]{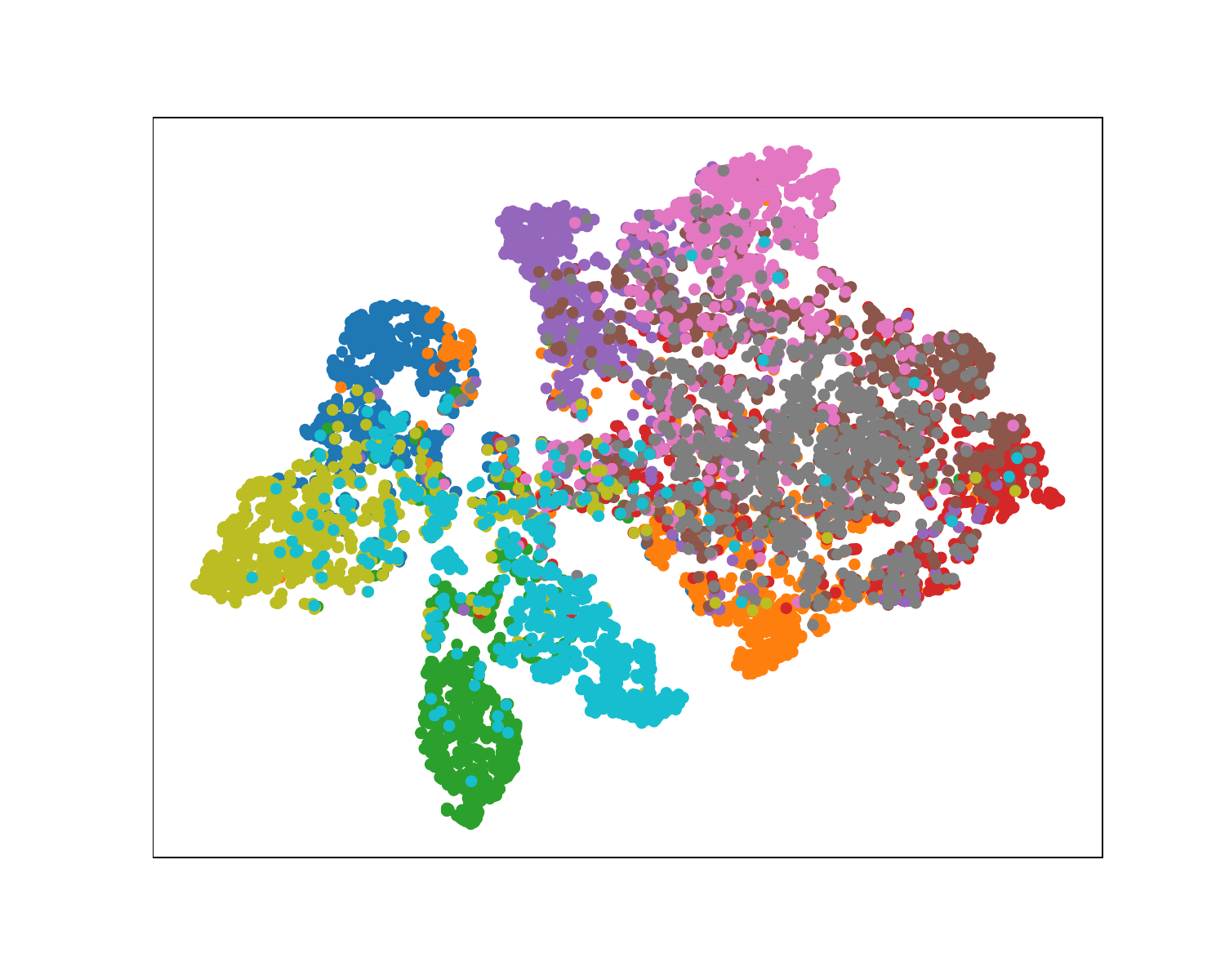}
        \caption{Ours (S)}
        \label{fig:sub12}
    \end{subfigure}

    \caption{The illustration shows the impact of representation learning by our methods and SimCLR with only 30 epochs fine-tuning on several datasets. C represents CIFAR10, M represents MNIST, and S is for STL10 datasets. Ours-H has trained models only Huber loss, and Ours-F represents FRD batch curation without Huber loss. Ours is a combination of Huber loss and FRD.}
    \label{fig:full}
\end{figure}

Figure \ref{fig:full} illustrates t-SNE plots for three distinct datasets. Notably, when trained with regularized loss, our method exhibits superior performance on the MNIST, CIFAR10 compared to SimCLR. However, the most optimal representation across all three datasets is achieved when the model employs both FRD and Huber loss. The relation between FRD and Huber loss is evident in their complementary roles; FRD eliminates weakly transformed views during gradient updates, while Huber loss mitigates the impact of false positives and false negatives within a batch. This dual approach contributes to the enhanced overall performance observed in the t-SNE plots.

Table~\ref{tbl:tab1} provides a comparison of our method against other contrastive learning approaches, comparing the number of parameters, the backbone network model and performance. All models are trained and tested on ImageNet. We note that the proposed method achieves superior performance with a significantly larger number of parameters, achieving an accuracy of 83.94$ \% $. This result supports our hypothesis that bad batches are a large hindrance to contrastive learning approaches. 

To verify whether the proposed approach allows for better performance in low-memory, low-data situations, we compared performance pre-training and testing on CIFAR10. 
Table~\ref{tbl:tab2} presents results obtained through varying epochs of training and batch sizes with the ResNet-50 backbone network. The proposed approach achieves the best top-1 accuracy of 99.67\%, with a batch size of only 128 and only 200 epochs of pre-training. It is worth noting that DINO \cite{dino} achieved comparable performance on CIFAR10 with 99.1$ \% $, but it is important to also note the larger number of epochs required (800 versus 200) and substantial parameterization of vision transformers which may pose memory consumption challenges during self-supervised training. 
The results also show that the achieved performance is comparable to the one achieved with a fully supervised training of the network.  

One question is whether it is the Huber regularization or the Fréchet batch curation that provides the main improvement. To elucidate this, table~\ref{tbl:tab3} provides an ablation study comparing regularization with $L_1$, $L_2$ or Huber loss, with and without curation, against standard SimCLR and PatchBatch~\cite{welle2021batch} another batch curation approach. The results show that both components lead to a large improvement in performance and that Huber loss regularization performs better than $L_1$ or $L_2$. This confirms that the Huber loss offers robustness to noise by modulating sensitivity to outliers through the parameter $\delta$, where $\delta = 1.0 $ proved effective. We assume that outliers might be false positives and negatives in our experiments. 
A key advantage of our proposed method lies in its ability to extract robust features from a small dataset. Table~\ref{tbl:tab3} highlights the performance of self-supervised batch curation methods pre-trained on CIFAR10 and tested on the same dataset. In contrast to other algorithms, our approach allows the model to acquire a resilient representation without necessitating extensive epoch training, and data augmentation configuration for each specific dataset. The Patch Curation algorithm \cite{welle2021batch}, utilizing Euclidean distance for batch selection, is surpassed by our method, particularly in cases where adjacent transformations construct patches without intersection. This implies that the Patch Curation algorithm may inadequately address false positives or negatives by relying solely on distance measurements within a batch. Conversely, our FRD batch curation method leverages pairwise data distribution analysis to effectively eliminate false negatives and false positives in representations.
Eventually, contrastive learning is often used for transfer learning, where a large, unlabelled dataset is available for pre-training, and a limited dataset is available for the downstream task. This raises the question of how well the proposed approach performs in this scenario. In table~\ref{tbl:tab4}, we compare the performance of our method against SimCLR, both pre-trained on ImageNet, against a fully supervised ResNet-50 trained on the downstream task. In table~\ref{tbl:tab4}, we present the fine-tuning results of our algorithms with an increased batch size (instead of 128). The table demonstrates that our method can learn representations even with larger batch sizes. When compared across various datasets, our algorithm consistently outperforms large-scale image datasets such as Flower102 and small-scale datasets like STL10 and CIFAR10. Notably, even in the case of the greyscale MNIST handwritten dataset, our method exhibits only marginally inferior performance compared to supervised results. This underscores the effectiveness of our approach in achieving superior performance across diverse datasets, demonstrating its potential to enhance generalization ability without the need for an extensive number of samples or large batch sizes. 
\begin{table}[ht]
  \centering
  \footnotesize
  \caption{Top-1 accuracy results on several representation learning methods with 200 epochs, 128 batch size, and non-linear projection head on ImageNet dataset.}
  \label{tbl:tab1}
  \begin{tabularx}{\linewidth}{| X | c | c | c |}
    \hline
    \textbf{Method} & \textbf{Backbone Network} &  \textbf{Param (M)}  & \textbf{Top-1}   \\ 
    \hline \hline
    SimCLR \cite{chen2020simple} &  ResNet-50  & - & 62.5  
    \\ \hline
    SimCLR \cite{chen2020simple}  & ResNet-50 $(2\times)$  & 94  & 74.2 
    \\
    \hline   SimCLR \cite{chen2020simple}  & ResNet-50 $(4\times)$  & 375  & 76.5 
    \\
    \hline   AMDIM \cite{amdim}  & Custom ResNet $(2\times)$  & 626  & 68.1 
    \\
     \hline   BYOL \cite{byol}  & ResNet-200 $(2\times)$  & 250  & 79.6 
    \\
     \hline   DINO \cite{dino}  & VIT-S   & 21  & 77.0
    \\
     \hline   DINO \cite{dino}  & VIT-B   & 85  & 78.2
    \\
     \hline   BINGO \cite{bingo}  & ResNet-34  & -  & 66.1
    \\
    \hline  SMoG \cite{smog} & ResNet-50 $(4\times)$ & 375  & 79.0
    \\
     \hline   VicRegL \cite{vicregL}  & ConvNext-S   & 50  & 75.9
    \\
     \hline   VicRegL \cite{vicregL}  & ConvNext-B   & 85  & 77.1
    \\
    \hline  Ours & ResNet-50 & 27.9 & \textbf{83.94}
    \\
    \hline
  \end{tabularx}
\end{table}
\begin{table}[ht]
\footnotesize
    \begin{center}
        \begin{tabular}{|c | c | c | c |} 
            \hline
            \textbf{Method} & \textbf{Batch Size} &\textbf{Epochs}   & \textbf{Top-1 Accuracy} \\ [0.5ex] 
            \hline
            \hline Supervised  & 128  & 200 & \textbf{99.87}
            \\
            \hline CaCo \cite{wang2023caco} & 128  & 200 &  92.6
            \\
         
            \hline ReSSL \cite{ressl} & 256  & 200  & 89.37
            \\
            \hline SimSiam \cite{SiameseIM} & 128  & 200 &  70.0
            \\  
            \hline SimCLR \cite{chen2020simple} & 128  & 200 & 62.5  
            \\
            \hline  SimSiam\cite{SiameseIM} & 512   & 800  &  91.8 
            \\
            \hline SimCLR \cite{chen2020simple}  & 256  & 200 &  87.5
            \\
            \hline   DINO \cite{dino}  & -  & 800  & 99.1
            \\
            \hline  Mixed Barlow Twins \cite{mixedbarlow} & 128   & 2000  &  92.58 
            \\
            \hline  Ours & 128 & 200 & \textbf{99.67} 
            \\
            \hline     
        \end{tabular}
    \caption{These are top-1 accuracy scores for the linear classifier testing on CIFAR10. }
    \label{tbl:tab2}
    \end{center}
\end{table}
\begin{table}[ht]
\tiny
    \begin{center}
        \begin{tabular}{|c | c | c | c | c |} 
            \hline
            \textbf{Method} & \textbf{FRD} &  \textbf{Data Augmentation} & \textbf{Loss}  & \textbf{Top-1 Accuracy}   \\ [0.5ex] 
            \hline  PatchBatch \cite{welle2021batch} & \xmark  & Intersection Crop &NT-Xent & 75.06
            \\
            \hline PatchBatch \cite{welle2021batch} &  \xmark & Adjacent Crop & NT-Xent & 76.31
            \\
            \hline SimCLR & \xmark  & Adjacent Crop & NT-Xent & 65.96
            \\
            \hline  Ours & \xmark  & Adjacent Crop & Reg. NT-Xent (Huber) & 73.99
            \\
            \hline  Ours & \xmark & Adjacent Crop & Reg. NT-Xent (L1) & 71.56
            \\
            \hline  Ours &  \xmark & Adjacent Crop & Reg. NT-Xent (L2) & 72.87
            \\
            \hline  Ours & \cmark  & Adjacent Crop & Reg. NT-Xent (Huber) & \textbf{83.92}
            \\
            \hline  Ours & \cmark & Adjacent Crop & Reg. NT-Xent (L1) & 70.87
            \\
            \hline  Ours &  \cmark & Adjacent Crop & Reg. NT-Xent (L2) & 68.49
            \\
            \hline     
        \end{tabular}
    \caption{These are top-1 accuracy scores for the k-Nearest Neighbour (k-NN) classifier, and the dataset represents self-supervised pertaining which is CIFAR10, testing on CIFAR10. }
    \label{tbl:tab3}
    \end{center}
\end{table}
\begin{table}[ht]
\scriptsize
    \begin{center}
   \begin{tabular}{|c | c | c | c | c | c |} 
            \hline
            \textbf{Method}  & \textbf{CIFAR100} & \textbf{STL10} & \textbf{Flower102}  & \textbf{Caltech101} & \textbf{MNIST} \\ [0.5ex] 
            \hline Supervised & \textbf{94.22}  & 83.26 & 93.34 & 86.07 & 95.56
            \\[1ex] 
            \hline SimCLR \cite{chen2020simple}  & 85.9 & - & 97.0 & \textbf{92.1}  & -
            \\[1ex] 
            \hline Ours  & 91.8 & \textbf{87.74} & \textbf{99.31} & 91.42 & \textbf{96.51}
            \\
            \hline
        \end{tabular}
    \caption{Comparison of transfer learning performance of the methods with several image datasets. These are top-1 accuracy scores for linear evaluation. SimCLR model uses ImageNet pre-trained ResNet-50 model and is fine-tuned with the datasets. Self-supervised learning batch size is 128, fine-tuning size is 1024 for CIFAR10, STL10, MNIST, and the batch size is 256 for Caltech101, and Flower102. For SimCLR paper results, the number of batch sizes is not given.  }
    \label{tbl:tab4}
    \end{center}
\end{table}

\begin{figure}[ht!]
    \centering
    \includegraphics[scale=0.5]{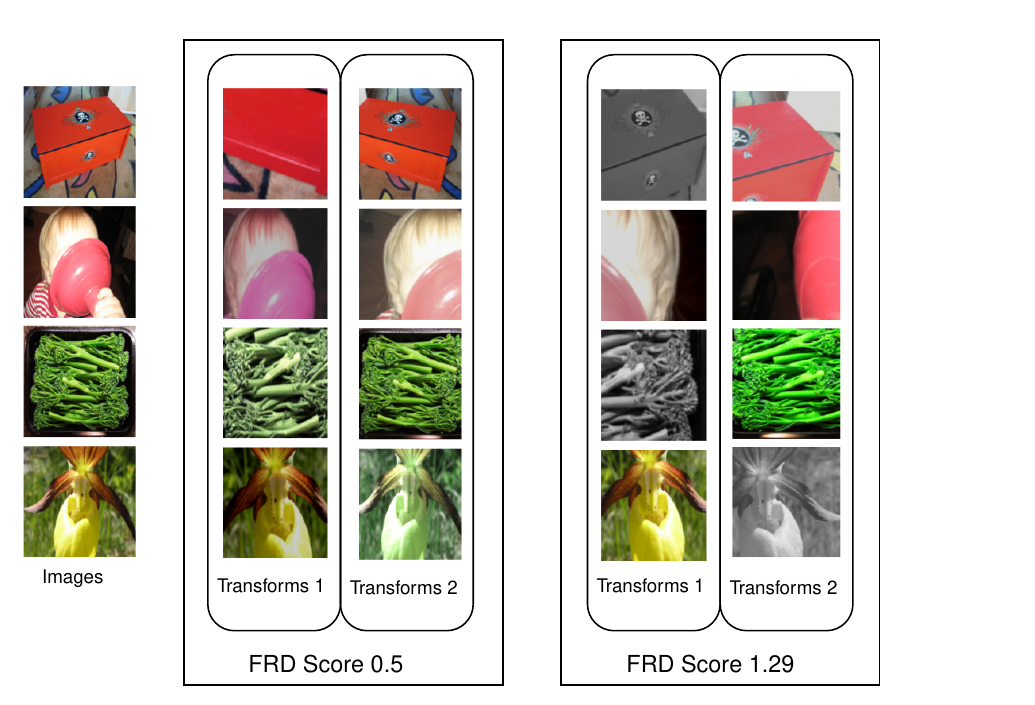}
    \caption{FRD score shows the quality of each batch in a training process. Higher FRD scores represent bad batches which have many false positives between batches. In this figure, transformed views with FRD score of 1.29 demonstrate the poor augmentation between sea images.}
    \label{fig:frd_scores}
\end{figure}
\section{Discussion}
This study introduces a novel framework statistically centred around evaluation batches, with the primary goal of facilitating unsupervised representation learning from datasets while addressing the impact of false positives on contrastive loss. Unlike several contemporary approaches that rely on pre-training models with extensive datasets like ImageNet, our framework seeks to overcome challenges associated with time-consuming iterations, limited dataset applicability in self-supervised learning, and false positives which are transformed views of raw input data.
A noteworthy contribution of our work to the existing literature lies in its focus on training self-supervised contrastive models efficiently across various datasets, without the need for data augmentation configurations, larger batch sizes or even large-scale datasets. This emphasis promotes reproducibility in diverse computing environments, eliminating the necessity for substantial memory requirements or prolonged training durations. Additionally, we underscore the crucial role of weakly transformed data augmentation in contrastive learning for effective representation learning.
While acknowledging the importance of data augmentation, we recognize its potential drawbacks, particularly in introducing false positives. Our approach addresses this challenge by leveraging (FRD) to quantify difference distributions. By doing so, we aim to mitigate the impact of weak transformed views, providing a more robust foundation for representation learning.
Our experimental results affirm the significance of batch curation algorithms in enhancing representation learning. The proposed framework demonstrates favourable performance across a diverse range of datasets, showcasing its adaptability and effectiveness. This not only contributes valuable insights to the ongoing discourse on representation learning but also establishes the practical utility of our approach in real-world applications.
\section{Conclusion}
Self-supervised learning has demonstrated notable success across various tasks, exhibiting comparable performance to supervised learning methodologies. Numerous state-of-the-art algorithms have employed randomly curated batches in their training processes. However, random batch selection and augmentation may introduce many misleading pairs, including false positives and false negatives within batches. Furthermore, the contrastive loss alone does not adequately address the challenges related to false positives and negatives in batches, instead relying on large batch size to mitigate the issue. 

In this study, we described a simple approach to address these limitations, mitigate the impact of false positives and quantify the distributions of similar and dissimilar pairs by evaluating the FRD score. The experiments demonstrate a markedly better convergence and overall performance of the learnt representations on a variety of downstream tasks, while necessitating smaller batch sizes and training epochs. 

In conclusion, our findings underscore the importance of careful batch selection in self-supervised contrastive learning, particularly because the occurrence of bad augmentation is unavoidable in practice on large varied datasets without a lengthy and expensive hyperparameter tuning of the data augmentation. We also note that the approach is generic, and could be extended to other contrastive learning approaches. 
\vspace{0.25cm} \\
\textbf{Acknowledgements:} \"Ozg\"u G\"oksu is supported by the Turkish Ministry of National Education.

{
\bibliographystyle{ieee_fullname}
\bibliography{egbib}
}

\end{document}